# Forecasting Transportation Network Speed Using Deep Capsule Networks with Nested LSTM Models

Xiaolei Ma [1,2], Yi Li [1], Zhiyong Cui [3*], Yinhai Wang [2,3*]

*1 School of Transportation Science and Engineering, Beijing Key Laboratory for Cooperative Vehicle Infrastructure System and Safety Control, Beihang University, Beijing 100191, China*
*2 Beijing Advanced Innovation Center for Big Data and Brain Computing, Beihang University, Beijing 100191, China*
*3 Department of Civil and Environmental Engineering, University of Washington, Seattle 98195, United States*

*\*Corresponding author*

**Abstract**

Accurate and reliable traffic forecasting for complicated transportation networks is of vital importance to modern transportation management. The complicated spatial dependencies of roadway links and the dynamic temporal patterns of traffic states make it particularly challenging. To address these challenges, we propose a new capsule network (CapsNet) to extract the spatial features of traffic networks and utilize a nested LSTM (NLSTM) structure to capture the hierarchical temporal dependencies in traffic sequence data. A framework for network-level traffic forecasting is also proposed by sequentially connecting CapsNet and NLSTM. On the basis of literature review, our study is the first to adopt CapsNet and NLSTM in the field of traffic forecasting. An experiment on a Beijing transportation network with 278 links shows that the proposed framework with the capability of capturing complicated spatiotemporal traffic patterns outperforms multiple state-of-the-art traffic forecasting baseline models. The superiority and feasibility of CapsNet and NLSTM are also demonstrated, respectively, by visualizing and quantitatively evaluating the experimental results.

## 1. Introduction

Traffic prediction has become crucial for individuals and public agencies due to the requirements of accurate travel time estimation and dynamic transportation management. Traffic prediction aims to forecast the future traffic states of connected roadway segments on the basis of historical traffic data within an underlying roadway network structure.

In the early stage, the majority of traffic prediction studies that focus on small-scale roadway networks are normally fulfilled based on statistical models with limited transportation data. In recent years, advanced data-driven machine learning methods have been widely adopted for network-wide traffic state prediction with the rapid development in traffic sensing technologies and computational power. Machine learning models have outperformed classical statistical models due to their capabilities of handling high-dimensional and complicated spatiotemporal data. However, the potential of machine learning models for traffic prediction has not been fully utilized until the rise of deep neural network (NN) models (also referred to as deep learning models) (Ma et al., 2015).

Deep learning models have achieved superior performance in traffic forecasting tasks compared with conventional machine learning models. With the utilization of fully connected NNs (Park and Rilett, 2010) for traffic prediction, many advanced and powerful deep learning models, such as deep belief networks (DBNs) (Huang et al., 2014), convolutional NNs (CNNs) (Ma et al., 2017), and recurrent NNs (RNNs) (Lint et al., 2002), have been applied to extract high-dimensional features of traffic states and have achieved good prediction performance. However, these models should be improved in terms of capturing the spatial and temporal dependencies in high-dimensional traffic data. Most of the existing studies on traffic prediction have modeled spatial dependencies with CNNs (Ma et al., 2017; Zhang et al., 2016) and captured temporal dependencies via RNNs (Ma et al., 2015; Cui et al., 2017). They (Yu et al., 2017; Li et al. 2018; Zhang et al., 2014) have also proposed models by combining CNNs and RNNs to fulfill this task. However, conventional CNNs and RNNs have their limitations when handling network-wide traffic data.

Conventional CNNs are appropriate for capturing spatial relationships in Euclidean space that are represented by two-dimensional (2D) matrices or images. On this basis, spatiotemporal traffic data learning using CNNs can be roughly categorized into two strategies. The first strategy (Ma et al., 2017) uses CNNs to learn spatiotemporal traffic data as a 2D matrix, in which the spatial and temporal dimensions are separately distributed in two directions. However, the actual structure of a complicated roadway network cannot be properly represented by a 2D matrix, and CNNs inevitably capture a certain amount of spurious spatial relationships. The second strategy (Zhang et al., 2016; Yu et al., 2017) employs CNNs to capture the spatial dependencies by projecting various traffic states to their corresponding physical roadway links using different colors and by processing a traffic network map as an image. In this way, the actual spatial features of the traffic network are learned. However, CNN-based feature extraction models still face challenges. First, the pooling operations in CNNs proactively discards substantial information; thus, critical correlations of traffic states between links may be lost. Then, the neurons in conventional CNNs are unsuitable in representing the various properties of a particular entity (Sabour et al., 2017), such as pose, deformation, albedo, hue, and texture. Given that the structure of a traffic network is fixed in a map-based image, the various colors of pixels located on roadways that represent traffic states can be considered as the textures or poses of the traffic network viewed from different perspectives. Thus, the CNN approach is insufficient to capture the relative spatial dependencies between colored pixels when these colors gradually change in a sequence of map-based images. In addition, the interdependencies between roadway links cannot be captured by CNNs for several specific complicated road network structures that contain viaducts, intersections, and side roads, as shown in **Fig.** 1.

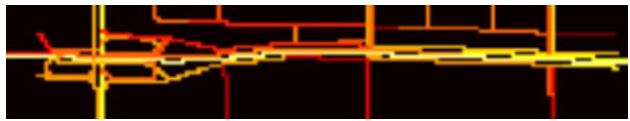

**Fig. 1.** Viaducts, intersections, and side roads in a traffic network

RNN and its variants, such as long short-term memory (LSTM) network (Hochreiter and Schmidhuber, 1997), are effective for capturing the temporal features of traffic states. Existing studies have used stacked LSTMs to enhance the short-term traffic prediction performance (Cui et al., 2017). Generally, traffic conditions are not only influenced by short-term historical information that is directly relevant to current traffic states but also by upper-level long-term traffic patterns with strong periodicity and regularity. However, long-term temporal dependencies under severe weather or disasters do not contribute considerable short-term information to traffic prediction and should be selectively captured. According to (Moniz and Krueger, 2018), a stacked LSTM or a single-layer LSTM cannot comprehensively characterize a temporal hierarchy.

To overcome the drawbacks of conventional CNNs and LSTMs, we propose a new capsule network (CapsNet) to extract the spatial features of traffic networks and utilize a newly proposed nested LSTM (NLSTM) structure to improve the performance of time-series learning. The CapsNet utilizes capsules in vector form rather than in scalar form as neurons in the NN. The direction and length of a capsule vector encode the state of high-level features and the detection probability of a feature, respectively. With the aid of capsules and a dynamic routing algorithm between them, the CapsNet considers slightly active features and largely addresses the existing problems in conventional CNNs. The NLSTM with the capability to access inner memories selectively in constructing temporal hierarchies is utilized to capture the hierarchical temporal dependencies in traffic data dynamically. The evaluation results show that the proposed framework with the combination of CapsNet and NLSTM outperforms multiple state-of-the-art traffic forecasting baseline models.

In summary, our main contributions are as follows.
1) We learn the traffic network as an image and propose a new CapsNet to capture the spatial dependencies between the roadway links and extract the high-level spatial features of network-level traffic states;
2) We utilize an NLSTM structure to capture the hierarchical temporal dependencies in traffic sequence data dynamically;
3) We propose a new framework for network-level traffic prediction by combining CapsNet and NLSTM. On the basis of literature review, our study is the first to adopt CapsNet and NLSTM in traffic forecasting;
4) The superiority and feasibility of CapsNet and NLSTM are demonstrated by visualizing and quantitatively evaluating the experimental results.

The remainder of this paper is organized as follows. Sections 2 and 3 presents the related works and our methodology, respectively. Section 4 shows the experimental data and results. Finally, Section 5 concludes the paper and open questions for future research.

2. **Literature review**

The approaches in the existing literature on short-term traffic prediction can be divided into two families, namely, statistical methods and artificial intelligence. Statistical methods, such as autoregressive integrated moving average (ARIMA) (Hamed et al., 1995), ARIMA variants (Voort et al., 1996; Williams and Hoel, 2003; Williams, 2001), Kalman filter (Okutani and Stephanedes, 1984; Guo et al., 2014), and exponential smoothing (Williams et al., 1998; Tan et al., 2009), have been investigated and applied to predict traffic flow parameters. In comparison with parametric statistical models, non-parametric machine learning models have more portability, higher accuracy, and are free of assumptions on data distribution (Davis and Nihan, 1991). For example, k-nearest neighbors (Zheng and Su, 2014; Cai et al., 2016) and support vector machines (Smola and Schölkopf, 2004; Wu et al., 2004), which are popular in the field of prediction, have been widely utilized to predict traffic speed and travel time.

However, as a component of artificial intelligence, machine learning methods may fail when addressing complicated high-dimensional data. In the early stage, several NN approaches, such as artificial NN (Huang and Ran, 2006), fuzzy NN (Yin et al., 2002), state-space NN (Van et al., 2005) and radial basis function NN (Messer et al., 1998; Zhu et al., 2014), were applied to predict traffic states. In recent years, considerable advanced and powerful deep learning models, such as DBNs (Huang et al., 2014), stacked autoencoders (Lv et al., 2015), CNNs (Ma et al., 2017; Liu et al., 2018), and RNNs (Ma et al., 2015; Lint et al., 2002; Cui et al., 2017), have been adopted in traffic forecasting. As a representative variant of RNNs, LSTM was first introduced in traffic prediction task and showed promising performance (Ma et al., 2015). On this basis, stacked bidirectional LSTMs are also adopted to enhance the short-term traffic prediction (Cui et al., 2017). Existing studies (Ma et al., 2017; Yu et al., 2017; Liu et al., 2018) have used CNNs in expanding the study areas to large-scale traffic networks, which are proven effective in computer vision and image recognition areas (Oquab et al., 2014), to extract spatial dependencies from traffic data to facilitate prediction performance. For example, a one-dimensional CNN has been used to capture the spatial features of traffic flow (Wu and Tan, 2016).

However, a common means of adopting CNNs for traffic forecasting is by processing 2D spatial–temporal data as images and by learning spatial–temporal dependencies from these images (Ma et al., 2017). To model the spatial–temporal relationships of traffic states effectively, hybrid CNN approaches that incorporate LSTM (Yu et al., 2017) and residual unit (Zhang et al., 2016) have been proposed for traffic prediction by learning the spatial features from the images converted from grid- or pixel-based traffic network maps. However, these approaches that adopt conventional CNNs still cannot handle the overlapping roadways caused by low-resolution images, separate spatially interlaced links in viaducts in 2D space, and accommodate the physical specialties of traffic networks.

We construct a new CapsNet to solve the limitations of CNN approach in extracting the spatial features of network-level traffic states and utilize a nested LSTM structure to improve the performance of time-series learning. The two methods are sequentially connected to build a deep learning architecture for the traffic prediction problem.

## 3. Methodology

### 3.1. Network representation

The traffic state of a roadway link in a road network is defined by the average speed of vehicles that travel on that link. The average speed of a link $a$ in period $t$ is calculated as follows:

$$V_{at} = \frac{\sum_{i=1}^{k} V_{it}}{k}, \tag{1}$$

X. Ma et al.

where $a \in (1, 2, \cdots, n)$, $k$ is the number of vehicles passing through the link during the time interval, and $V_{it}$ represents the average speed of each vehicle.

To learn traffic as images, the average speed of each link is projected in the road network combined with a GIS map to establish the spatial correspondence between the links and traffic states. As shown in **Fig. 2**, the calculated speed on each link is visualized with different colors. We then convert the color image to a single-channel grayscale image.

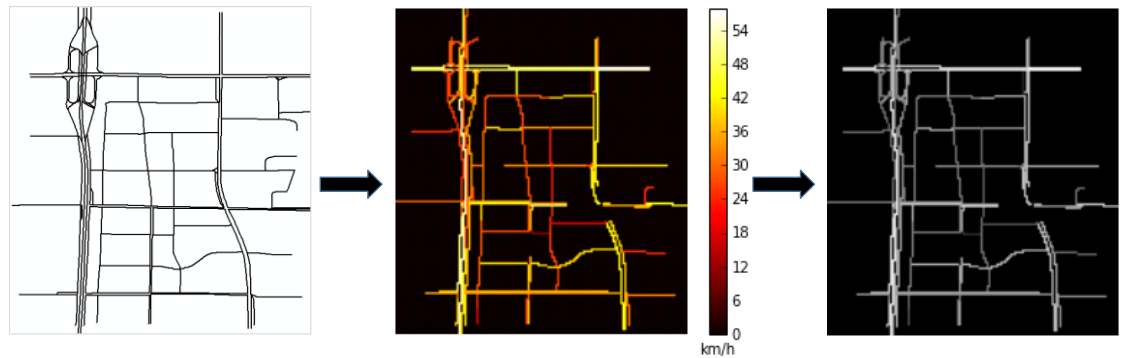

**Fig. 2.** Transportation network representation

The traffic states of the road network are characterized by matrix images through a gridding process. The road network is divided into multiple grids with a certain spatial latitude and longitude range. The schematic of processing is shown in **Fig. 3**, in which a small part of the road network is used as an example. First, the road network is segmented by grids with a size of 0.0001° × 0.0001° (latitude and longitude). Subsequently, the value of each grid is determined on the basis of the speed of links using the following criteria: if no link passes through the grid area, then the value is zero; if only one link passes through the grid area, then the value is the speed of this link; if multiple links occupay the same grid area, then their average speed is assigned to the grid.

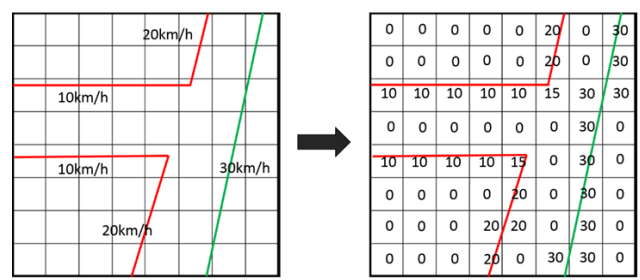

**Fig. 3.** Schematic of the gridding process

On the basis of the above process, each grid is taken as a pixel with one channel, in which its value is the projected velocity value. Sequences of images are generated as data samples, and the time interval in these sequences is 2 min. These images not only represent the traffic state but also contain the spatial structure of the road network and the relative topology among different links.

*3.2. Spatial features captured by CapsNet*

In the aforementioned review, the CNN approach has shown promising results in capturing the spatial relationships among the links in urban road networks, where a congestion occurring on a far-side road segment may also influence the near-side traffic condition. However, this method has several important drawbacks. First, the CNN approach extracts the spatial dependencies on many distant links by using successive convolutional layers or max pooling, where valuable information is lost. Second, the CNN approach cannot effectively distinguish between two links that are not spatially connected for several complicated road structures, such as viaducts. Third, the CNN approach cannot sufficiently handle the overlapping areas due to the low resolution in the grid-based image process. Thus, in this study, a CapsNet is utilized to solve the limitations of the CNN approach in extracting the spatial features of network-level traffic states.

A CapsNet is a new type of NN structure that is characterized by the use of "capsules" in a vector form rather than traditional scalar forms of neurons. In extracting the local features in images, all important information about the state of the features that the capsules detect is encapsulated in the vector form. Particularly, the length of an output vector encodes the detection probability of a feature. The direction of the vector encodes the state of the features, such as rotation angle, direction, and size. The CapsNet inherently can detect multiple objects. Therefore, the CapsNet can effectively distinguish the spatial interlaced links and overlapping regions on several complicated road structures and low-resolution problem in traffic images by using such state of the features implied in the output vectors. In addition, the CapsNet can retain all the extracted local features by replacing the pooling operation with a dynamic routing operation between the capsules and thus avoid the problem of missing several spatial relationships among the links.

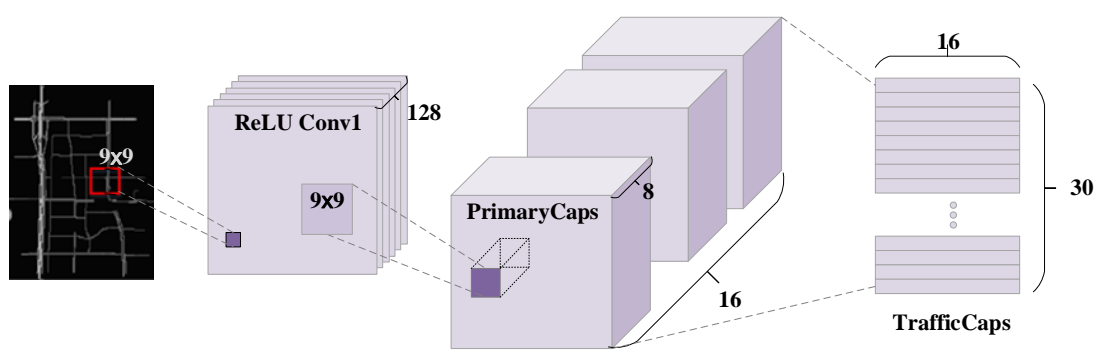

**Fig. 4.** Layers of CapsNet

In this study, the CapsNet model is composed of two convolutional layers and a fully connected layer (called TrafficCaps), as shown in **Fig. 4**. For the input images that represent the traffic state of the road network, the first convolutional layer is used to extract the spatial relations between the adjacent links, that is, the local features of traffic states. The second convolution layer is then utilized in the primary capsule layer (named PrimaryCaps), and the "neurons" that use a single scalar output are converted to primary capsules in the vector form with a dimension of 8. Finally, the TrafficCaps is employed to capture the spatial relationship between the local features implied in all primary capsules and to output the features to a set of advanced capsules with a dimension of 16. The details of each part are subsequently explained.

In CNNs, high-level neurons receive input scalars from low-level neurons through weighting operations and activation functions, and the weights are learned by backpropagation (Lecun et al., 1990). By contrast, the weighting operations, activation functions, and learning method of weights between primary and advanced capsules are different because the capsules are in the vector form in the CapsNet.

The first convolution layer is the same as the convolutional layer in CNN by using ReLU as the activation function. The latter two layers use a novel nonlinear "squashing" activation function for the vector form of capsules, as shown as follows:

$$v_j = \frac{\|s_j\|^2}{1+\|s_j\|^2} \frac{s_j}{\|s_j\|}, \qquad (2)$$

where $v_j$ is the output vector of capsule $j$, and $s_j$ is the input vector. The squashing operation ensures that the short vectors shrink to approximately zero length and long vectors shrink to a length slightly below 1. Thus, the length of the output vector of a capsule can represent the probability of the existence of the extracted local features.

To obtain the spatial relationship between the local features of network-level traffic state extracted by the PrimaryCaps layer and advanced features, an affine transformation is performed by multiplying the local features with a weight matrix $W_{ij}$.

$$\hat{u}_{j|i} = W_{ij} u_i, \qquad (3)$$

where $u_i$ is the local features extracted by a primary capsule $i$, and $\hat{u}_{j|i}$ is the input vector associated with an advanced capsule $j$.

For the TrafficCaps, input $s_j$ to an advanced capsule $j$ is the weighted sum over all input vectors $\hat{u}_{j|i}$ from the primary capsules in the layer.

$$s_j = \sum_i c_{ij} \hat{u}_{j|i}, \qquad (4)$$

where weights $c_{ij}$ are the coupling coefficients that determined by an iterative dynamic routing algorithm (Sabour et al., 2017). The essence of the dynamic routing algorithm is to find a part of primary capsules that is highly correlated to the advanced capsules, that is, to determine the local features with high probability to be associated with the high-level feature. This process represents the capability of the model to explore the spatial relationships among the distant links. For example, the dynamic routing algorithm will associate advanced capsule $j$ with a set of primary capsules that contains the local features that affect congestion when it represents a severe congestion at a viaduct. The specific process of the dynamic routing algorithm is described as follows.

1) For each primary capsule $i$ in the PrimaryCaps layer, the coupling coefficients $c_{ij}$ with all the advanced capsules $j$ are summed to 1 by using a SoftMax function:

$$c_{ij} = \frac{\exp(b_{ij})}{\sum_k \exp(b_{ik})}, \qquad (5)$$

where routing logit $b_{ij}$ is the log prior probability that capsule $i$ should be coupled to capsule $j$, and output $c_{ij}$ represents the normalized probability that primary capsule $i$ is associated with advanced capsule $j$. In the first iteration, the initial value of routing logit $b_{ij}$ is set to zero in which the probabilities of the primary capsule accepted by each advanced capsule are equal.

2) After all the weights $c_{ij}$ are calculated for all the primary capsules, each advanced capsule $j$ of the TrafficCaps is weighted by using Equation (4).

3) In this step, all the capsules from the last step are activated by the squashing nonlinear function, as shown in Equation (2). In this process, the direction of the vector is preserved in output $v_j$ and its length is enforced to be less than 1, which corresponds to the detection probability of high-level features.

4) In the iteration process, the initial coupling coefficients are iteratively refined by updating $b_{ij}$ on the basis of the following rule:

$$b_{ij} = b_{ij} + \hat{u}_{j|i} \cdot v_j. \qquad (6)$$

Routing logit $b_{ij}$ is updated by using the dot product of the input to capsule $j$ and its output. In the field of mathematics, the dot product becomes large for similar vectors. Therefore, the corresponding routing logit increases when the input and output are similar; thus, the primary capsule is coupled to the advanced capsule with a similar output. This process represents the association of local features with the high-level feature.

5) The algorithms from Steps 1–4 are repeated several times to obtain the optimal routing weights. The dynamic routing algorithm is easy to be optimized, and experiments show that the CapsNet model can be optimized by iterating three times on an MNIST dataset (Lecun and Cortes, 2010).

Overall, a set of vectors is generated to express the spatial features of network-level traffic state by applying the CapsNet model on the input traffic images for subsequent operations in the next step.

*3.3. Long short-term temporal features captured by NLSTM*

The traffic state normally has strong time evolution patterns and long-term dependencies, and a congestion state may last for several hours. LSTMs, which introduce memory units to optionally decide information through different cell states, have achieved promising learning capability of long-term time series (Hochreiter and Schmidhuber, 1997). In this study, a novel nested architecture of LSTMs, which is evaluated to outperform stacked and single-layer LSTMs on various character-level language modeling tasks (Moniz and Krueger, 2018), is used to capture the temporal features of traffic state. In stacked LSTMs (Cui et al., 2017), all the information extracted from a low LSTM must be inputted to the subsequent high-level LSTM layer and must be filtered again.

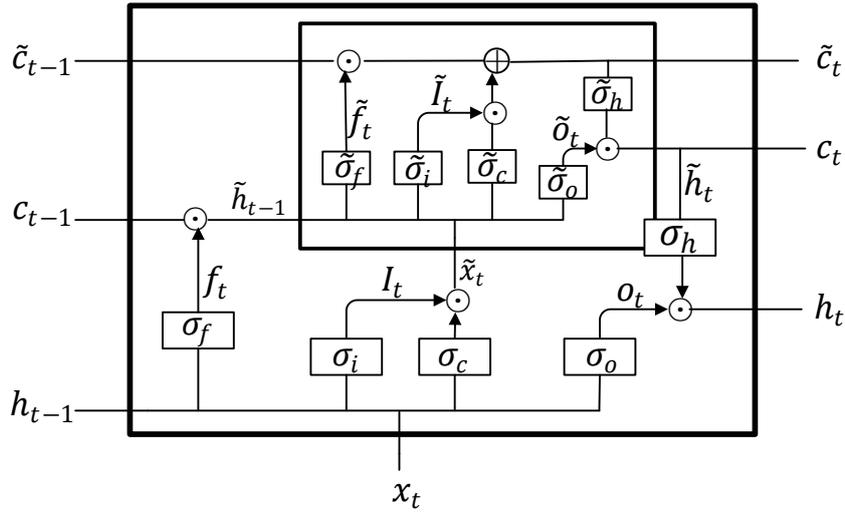

**Fig. 5.** NLSTM architecture

NLSTMs add depth to LSTMs by nesting rather than stacking. As shown in **Fig. 5**, the value of a memory cell in an NLSTM is computed by using an LSTM structure, which acts as an internal unit that has its own inner memory cells. The long-term information learned by the internal unit can be selectively read and written by using the standard LSTM gates. This process enables the inner memories to remember and process traffic events on long time scales, especially when these events are irrelevant to the immediate present. Such selective access to inner memories in NLSTM exhibits a stable and efficient performance in capturing the long-term dependencies of traffic states.

The equations that update the cell state and gates in an internal LSTM unit are similar to the standard LSTM unit, as shown as follows (the parameters with superscript ~ belong to the internal LSTM unit):

$$\tilde{I}_t = \tilde{\sigma}_i(\tilde{x}_t \widetilde{W}_{xi} + \tilde{h}_{t-1} \widetilde{W}_{hi} + \tilde{b}_i), \tag{7}$$

$$\tilde{f}_t = \tilde{\sigma}_f(\tilde{x}_t \widetilde{W}_{xf} + \tilde{h}_{t-1} \widetilde{W}_{hf} + \tilde{b}_f), \tag{8}$$

$$\tilde{c}_t = \tilde{f}_t \odot \tilde{c}_{t-1} + \tilde{I}_t \odot \tilde{\sigma}_c(\tilde{x}_t \widetilde{W}_{xc} + \tilde{h}_{t-1} \widetilde{W}_{hc} + \tilde{b}_c, \tag{9}$$

$$\tilde{o}_t = \tilde{\sigma}_o(\tilde{x}_t \widetilde{W}_{xo} + \tilde{h}_{t-1} \widetilde{W}_{ho} + \tilde{b}_o), \tag{10}$$

$$\tilde{h}_t = \tilde{o}_t \odot \tilde{\sigma}_h(\tilde{c}_t), \tag{11}$$

where $\tilde{x}_t$ and $\tilde{h}_{t-1}$ are the inputs of the internal LSTM unit and are calculated based on the parameters of the external unit, as shown as follows:

$$\tilde{x}_t = I_t \odot \sigma_c(x_t W_{xc} + h_{t-1} W_{hc} + b_c), \tag{12}$$

$$\tilde{h}_{t-1} = f_t \odot c_{t-1}, \tag{13}$$

where $\tilde{I}_t$, $\tilde{f}_t$, and $\tilde{o}_t$ are the three states of the gates; $\tilde{c}_t$ is the cell input state; $\widetilde{W}_{xi}$, $\widetilde{W}_{xf}$, $\widetilde{W}_{xo}$, and $\widetilde{W}_{xc}$ are the weight matrices connecting $\tilde{x}_t$ to the three gates and cell input; $\widetilde{W}_{hi}$, $\widetilde{W}_{hf}$, $\widetilde{W}_{ho}$, and $\widetilde{W}_{hc}$ are the weight matrices that connect $\tilde{h}_{t-1}$ to the three gates and cell input; $\tilde{b}_i$, $\tilde{b}_f$, $\tilde{b}_o$, and $\tilde{b}_c$ are the biases of the three gates and cell input; $\sigma$ represents the sigmoid function; and $\odot$ represents the scalar product of two vectors.

For the external LSTM unit, only the cell state update rule is changed to the output of the internal LSTM.

$$c_t = \tilde{h}_t \tag{14}$$

In this study, the temporal features of the traffic state are iteratively calculated by using the NLSTM model for traffic prediction.

### 3.4 Framework

The spatiotemporal features of the traffic state can be learned by the CapsNet and the NLSTM. We sequentially integrate CapsNet and NLSTM to forecast the future traffic states. The outputs of the CapsNet are spread in one vector and are passed to the NLSTM as the input, as shown as follows:

$$x_t = \{v_j^t\}_{j=1}^p, \tag{15}$$

where $v_j^t$ is the output vector of advanced capsule $j$ at timestamp $t$, and $p$ is the number of advanced capsules. At the end of the model, a fully connected layer is added after the NLSTM model to obtain the predictions of the traffic states of all links. The predicted speed is calculated as follows:

$$y^{t+1} = w \times h_t + b, \tag{16}$$

where $h_t$ is the output of the NLSTM; and $w$ and $b$ represent the weight and bias between the hidden layer and the fully connected layer, respectively. $y^{t+1}$ is the final output vector with the size of the number of links.

The entire prediction model of the network-level traffic state is shown in **Fig. 6**. The model is trained from end to end, and multi-step predictions are conducted based on the historical data of several steps.

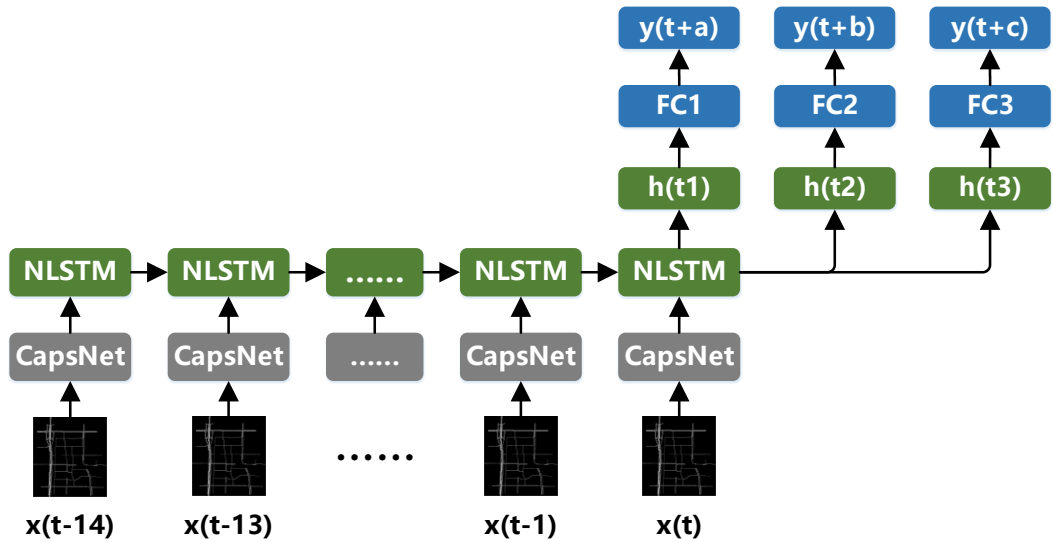

**Fig. 6.** Architecture of the prediction model (FC = Fully connected layers)

## 4. Empirical study

*4.1. Data description*

The traffic data used in the experiment were collected from the GPS devices mounted on floating vehicles. The time interval of data uploading approximately ranged from 10 s to 1 min, which depends on the sampling resolutions of GPS devices. Generally, narrow intervals may generate invalid data with average speed of zero and affect the traffic prediction performance. Thus, the time interval was aggregated to 2 min to capture the traffic state variations of the road network in this study accurately.

The dataset was divided into two subsets for training and testing to validate the effectiveness of the proposed prediction model. The training set was collected from June 1–31, 2015, and the test set was collected from August 1–14, 2015. The evaluated roadway network in this study encompasses 278 links, which include arterial roads, interchanges, and intersections, that are located between the Second and Third Ring Roads in Beijing. After the gridding process, the traffic states of the network are represented by an image with a size of 164 × 148 for 2 min.

In the experiment, the time lag of the input sequence was set to 15, which indicated that the traffic states of the previous 30 min were used as the input of the proposed model. The 30 min historical traffic speeds were used to predict the following 2, 10, and 20 min traffic speeds, which corresponded to the number of the time lags (a, b, c) = (1, 5, 10) in Fig. 6.

*4.2. Implementation*

*4.2.1. Hardware*

The deep learning model was implemented by using Python Keras (Chollet, 2018) and was executed on a server with 8 NVIDIA GeForce Titan X GPUs (12 GB RAM).

*4.2.2. Model parameters*

The details of our CapsNet+NLSTM deep learning model are shown in **Table 1**. The input has three dimensions, where the first two dimensions represent the resolution of the input image, and the last dimension indicates the amount of channel of the input image. The model was trained using the optimizer RMSprop (Tieleman and Hinton, 2012). The learning rate was set to 0.001 with 0.5 decay parameter for every 20 epochs, and the batch size was set to 32. A dropout layer was applied to prevent the problem of overfitting (Srivastava et al., 2014), and a fivefold cross-validation was used to determine the parameters of our deep learning model. In the cross-validation, the train set was divided into five subsets. Four subsets were used for training, and the remaining subset was used for validation. The optimal model has the lowest average prediction error in all validation datasets.

**Table 1** Model structure of CapsNet+NLSTM

| Name of layers | Parameters | Output | Parameter scale |
| --- | --- | --- | --- |
| Input | | $164 \times 148 \times 1$ | 0 |
| Convolution | Kernel size = $9 \times 9$<br>Channels = 128<br>Stride = 2 | $78 \times 70 \times 128$ | 10,496 |
| PrimaryCaps<br>(Convolution) | Kernel size = $9 \times 9$<br>Channels = 128<br>Stride = 4 | $18 \times 16 \times 128$ | 1,327,232 |
| Reshape | Capsule dimension = 8 | $4,608 \times 8$ | 0 |
| TrafficCaps<br>(Fully connected) | Advanced capsule = 30<br>Capsule dimension = 16 | $30 \times 16$ | 17,694,720 |
| (Flattened) | | 480 | 0 |
| NLSTM | Hidden unit = 800 | 800 | 9,222,400 |
| Dropout | 0.2 | 800 | 0 |
| Fully connected | | 278 | 222,678 |
| Total parameters | | | 28,477,526 |

*4.2.3. Baseline models*

We compared the proposed model with five baseline deep NN models, namely, LSTMs (Hochreiter and Schmidhuber, 1997), NLSTM (Moniz and Krueger, 2018), DCNNs (Ma et al., 2017), CapsNet (Sabour et al., 2017), and CNN+LSTMs (Yu et al., 2017), to evaluate its prediction performance. The details of the CNN+LSTM baseline model is shown in **Table 2**, and the total parameters are approximately half of the CapsNet+NLSTM model. For the LSTM, NLSTM, DCNNs, and CapsNet models, their structures were the same as the part of the two combined models. The LSTM model was constructed by stacking two standard LSTMs with 800 hidden units. The NLSTM model was a nested structure with the same 800 hidden units. For the single model of DCNNs and CapsNet, a flattened layer was added on the fully connected layer to integrate the outputs of 15 time steps into one vector for prediction.

**Table 2** Model structure of CNN+LSTMs

| Name of layers | Parameters | Output | Parameter scale |
|---|---|---|---|
| Input | | 164 × 148 × 1 | 0 |
| Convolution1 | Filter (3 × 3 × 16) | 82 × 74 × 16 | 160 |
| Pooling1 | Pooling (2 × 2) | | |
| Convolution2 | Filter (3 × 3 × 32) | 41 × 37 × 32 | 4,640 |
| Pooling2 | Pooling (2 × 2) | | |
| Convolution3 | Filter (3 × 3 × 64) | 21 × 19 × 64 | 18,496 |
| Pooling3 | Pooling (2 × 2) | | |
| Convolution4 | Filter (3 × 3 × 128) | 11 × 10 × 128 | 73,856 |
| Pooling4 | Pooling (2 × 2) | | |
| Flattened | | 14,080 | 0 |
| LSTM1 | Hidden unit = 800 | 800 | 47,619,200 |
| LSTM2 | Hidden unit = 800 | 800 | 5,123,200 |
| Fully connected | | 278 | 222,678 |
| Total parameters | | | 53,062,230 |

*4.2.4 Evaluation metrics*

The deep learning models in this study were evaluated by using two commonly used metrics in traffic forecasting, namely, mean squared error (MSE) and mean absolute percentage error (MAPE), which can be expressed as follows:

$$MSE = \frac{1}{n}\sum_{i=1}^{N}(\hat{y}_i - y_i)^2, \quad (17)$$

$$MAPE = \frac{1}{n}\sum_{i=1}^{N}\left(\frac{\hat{y}_i - y_i}{\hat{y}_i}\right), \quad (18)$$

where $\hat{y}_i$ is the prediction result of sample $i$, and $y_i$ is the ground truth of the corresponding traffic speed.

*4.3. Experimental results*

*4.3.1. Comparison*

In this section, we compared our CapsNet+NLSTM model with the other five baseline models and evaluated the prediction results by using the MSE and MAPE metrics. **Table 3** shows the comparison of different models for 1, 5, and 10 step-ahead predictions.

**Table 3** Comparison among different methods

| Time steps | 2 min | | 10 min | | 20 min | |
|---|---|---|---|---|---|---|
| Metrics | MSE | MAPE | MSE | MAPE | MSE | MAPE |
| LSTMs | 41.67 | 0.2158 | 44.67 | 0.2255 | 48.11 | 0.2273 |
| NLSTM | 39.55 | 0.2067 | 44.49 | 0.2229 | 47.32 | 0.2246 |
| DCNNs | 42.94 | 0.2131 | 47.14 | 0.2367 | 51.38 | 0.2384 |
| CapsNet | 35.80 | 0.1891 | 42.53 | 0.2205 | 47.08 | 0.2308 |
| CNN+LSTMs | 36.57 | 0.2051 | 43.10 | 0.2181 | 45.90 | 0.2258 |
| **CapsNet+NLSTM** | **31.04** | **0.1757** | **39.29** | **0.2071** | **42.88** | **0.2183** |

Among the three prediction steps, our CapsNet+NLSTM model yields the most accurate results in terms of MSE and MAPE. The average MSE values for CNN+LSTMs decrease by 15%, 8.8% and 6.6%. The CapsNet model performs better than the CNN model with 16.6%, 9.8%, and 8.4% lower MSE. This finding indicates that CapsNet shows stronger capability compared with CNNs in terms of the extraction of spatial features. For long-term temporal features, the prediction error increases with the prediction horizon, and the gap between the NLSTM and CapsNet+NLSTM models becomes small. This phenomenon indicates that the temporal features play an important role for the traffic prediction with the increase of prediction step size. Notably, the proposed model utilizes less parameters compared with the CNN+LSTM model and achieves more accurate results, which indicates that the CapsNet+NLSTM model achieves superior performance in predicting traffic states and shows a promising potential to be utilized.

*4.3.2. Evaluation of CapsNet*

We visualized and compared the prediction results of the CapsNet+NLSTM and CNN+LSTM models to evaluate the superior capability of CapsNet in extracting the spatial features of traffic states implied in the traffic images of complicated road networks. As shown in **Fig. 7**, we highlighted the links with a mean absolute error of speed more than 2 km/h, which were considered inaccurate predictions. In comparison with the CNN+LSTM prediction results in Fig. 7 (A) with 83 inaccurate links, the accuracy of CapsNet+NLSTM model exhibits an outstanding improvement with only 17 links highlighted. Furthermore, the inaccurate prediction results in Fig. 7 (A) are mainly concentrated in the viaducts and low- resolution areas with links that are tightly arranged. This condition verifies the poor performance of CNN approaches in distinguishing the links that are not spatially connected and processing the overlapping areas in traffic images, as previously explained. By contrast, the CapsNet-based model can achieve accurate predictions under these conditions.

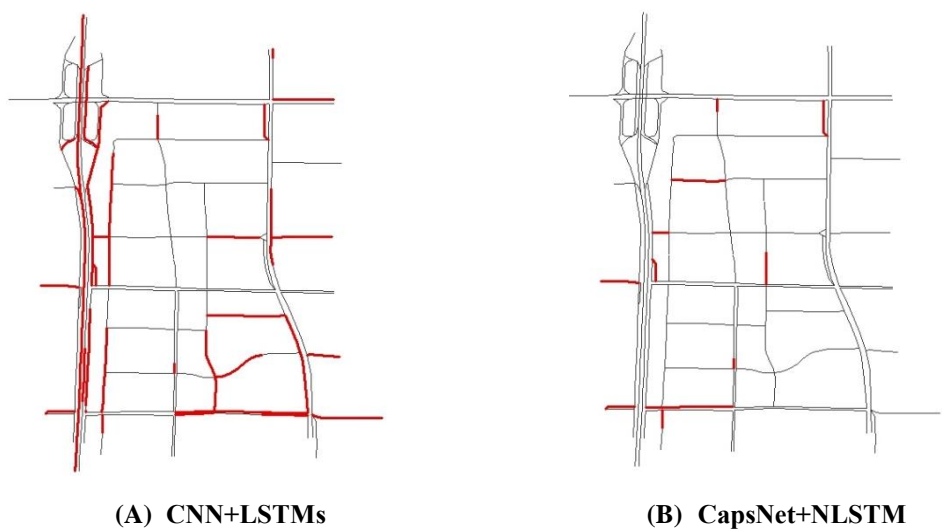

(A) CNN+LSTMs      (B) CapsNet+NLSTM

**Fig. 7.** Visualization of prediction results (The links with inaccurate predictions whose mean absolute errors are more than 2 km/h are marked in red.)

*4.3.3. Evaluation of NLSTM*

Moreover, we compared the NLSTM with LSTM in a long time lag with 40-min historical traffic speeds as inputs to evaluate the performance of NLSTM in learning long-term features. The results are shown in **Table 4**.

Table 4 Comparison of NLSTM and LSTMs

| Category | Prediction steps | 2 min | | 10 min | | 20 min | |
|---|---|---|---|---|---|---|---|
| | Time lags | 30 min | 40 min | 30 min | 40 min | 30 min | 40 min |
| Loss | LSTMs | 41.67 | 43.09 | 44.67 | 45.12 | 48.11 | 47.32 |
| (MSE) | NLSTM | 39.55 | 39.41 | 44.49 | 44.65 | 47.32 | 46.96 |
| Efficiency | LSTMs | 42 | 47 | 34 | 46 | 34 | 45 |
| (s) | NLSTM | 19 | 23 | 18 | 23 | 20 | 22 |

In comparison with LSTMs, the MSE values of NLSTM fluctuate more slightly when using long-term historical data in predicting multistep traffic speeds, especially in the 2-min interval short-term prediction. In terms of the number of parameters, the stacked and nested structures have the same scale of 8.8 M, and the NLSTM algorithm consumes small time in performing the prediction. These comparison results indicate that the NLSTM exhibits a stable and efficient performance in learning a long time series, which is the same as we expected.

Overall, the CapsNet outperforms the CNNs in capturing the spatial features of traffic states, and the NLSTM performs better in terms of stability and efficiency compared with the stacked LSTMs. These results demonstrate the superiority of our CapsNet+NLSTM model over the state-of-the-art deep learning algorithms, which shows promising potential in forecasting the traffic states of large-scale urban road networks.

## 5. Conclusions

Traffic prediction remarkably influences the overall performance of traffic management and control systems. In this study, a CapsNet+NLSTM approach is presented to address the important drawbacks of statistical models and machine learning methods in handling the complex spatial relationships among the links when performing network-level traffic state prediction. We use the traffic roadway network as an image to capture the spatial structure of the road network and the relative topology among the different links. Many spatial relationships among the links are preserved, and considerable spatial features of the network are encapsulated in the vector form of capsules, such as position, direction, length, and travel speed of the road segment, by using the new CapsNet rather than conventional CNNs. The incorporated NLSTM model can achieve a stable performance in time-series prediction compared with the traditional stacked structure of LSTM. The experimental results indicate that the CapsNet+NLSTM model outperforms other baseline models.

The major contributions of this study are summarized as follows. (1) A new CapsNet is developed to extract the comprehensive spatial features of roadway networks. (2) An NLSTM model is sequentially incorporated to capture the hierarchical temporal dependencies of traffic states. (3) The proposed model with the capability of capturing complicated spatiotemporal traffic patterns achieves the best prediction performance compared with the baseline models. (4) The visualized prediction results display the proposed model's promising capability in handling complicated road networks that contain interlaced and compact links, such as viaducts and side roads.

Several potential extensions in this research are considered. For example, the dynamic routing

X. Ma et al.

algorithm between capsules will be improved. Specifically, the prediction accuracy and model efficiency should be increased because the dynamic routing algorithm is the core component of the CapsNet. Furthermore, the interpretation of the learned spatiotemporal features captured by the CapsNet and NLSTM will be investigated in the future.

**Acknowledgments**

This paper is supported by the National Natural Science Foundation of China (U1564212 and 61773036), Beijing Natural Science Foundation (9172011), and Young Elite Scientist Sponsorship Program by the China Association for Science and Technology (2016QNRC001). The authors also thank Future Transportation and Urban Computing Joint Lab, AutoNavi company for data support.

**References**

Amin, S. M., Rodin, E. Y., Liu, A. P., Rink, K., García-Ortiz, A., 1998. Traffic prediction and management via RBF neural nets and semantic control. Computer‐Aided Civil and Infrastructure Engineering, 13(5), 315-327.

Cai, P., Wang, Y., Lu, G., Chen, P., Ding, C., Sun, J., 2016. A spatiotemporal correlative k-nearest neighbor model for short-term traffic multistep forecasting. Transportation Research Part C, 62, 21-34.

Chollet, F., 2018. Keras documentation. https://keras.io/.

Cui, Z., Ke, R., Wang, Y., 2017. Deep bidirectional and unidirectional LSTM recurrent neural network for network-wide traffic speed prediction. 6th International Workshop on Urban Computing (UrbComp 2017).

Davis, G. A., Nihan, N. L., 1991. Nonparametric regression and short‐term freeway traffic forecasting. Journal of Transportation Engineering, 117(2), 178-188.

Guo, J., Huang, W., Williams, B. M., 2014. Adaptive Kalman filter approach for stochastic short-term traffic flow rate prediction and uncertainty quantification. Transportation Research Part C, 43, 50-64.

Hamed, M. M., Al-Masaeid, H. R., Said, Z. M. B., 1995. Short-term prediction of traffic volume in urban arterials. Journal of Transportation Engineering, 121(3), 249-254.

Hochreiter, S., Schmidhuber, J., 1997. Long short-term memory. Neural Computation, 9(8), 1735-1780.

Huang, S. H., Ran, B. (2006). An application of neural network on traffic speed prediction under adverse weather condition. Transportation Research Board Annual Meeting.

Huang, W., Song, G., Hong, H., Xie, K., 2014. Deep architecture for traffic flow prediction: deep belief networks with multitask learning. IEEE Transactions on Intelligent Transportation Systems, 15(5), 2191-2201.

Krizhevsky, A., Sutskever, I., Hinton, G. E., 2012. ImageNet classification with deep convolutional neural networks. International Conference on Neural Information Processing Systems (Vol.60, pp.1097-1105). Curran Associates Inc.

Lecun, Y., Boser, B., Denker, J. S., Howard, R. E., Habbard, W., Jackel, L. D., et al., 1990. Handwritten digit recognition with a back-propagation network. Advances in Neural Information Processing Systems, 2(2), 396--404.

Lecun, Y., Cortes, C., 2010. The mnist database of handwritten digits.

http://yann.lecun.com/exdb/mnist/.

Li, Y., Yu, R., Shahabi, C., Liu, Y., 2018. Diffusion convolutional recurrent neural network: data-driven traffic forecasting. International Conference on Learning Representations (ICLR '18)

Lint, J. W. C. V., Hooqendoorn, S. P., Zuvlen, H. J. V., 2002. Freeway travel time prediction with state-space neural networks: modeling state-space dynamics with recurrent neural networks. Transportation Research Record Journal of the Transportation Research Board, 1811(1), 347–369.

Liu, Q., Wang, B., Zhu, Y., 2018. Short-Term Traffic Speed Forecasting Based on Attention Convolutional Neural Network for Arterials. Computer-Aided Civil and Infrastructure Engineering, doi:10.1111/mice.12417.

Lv, Y., Duan, Y., Kang, W., Li, Z., Wang, F. Y., 2015. Traffic flow prediction with big data: a deep learning approach. IEEE Transactions on Intelligent Transportation Systems, 16(2), 865-873.

Ma, X., Dai, Z., He, Z., Ma, J., Wang, Y., Wang, Y., 2017. Learning traffic as images: a deep convolutional neural network for large-scale transportation network speed prediction. Sensors, 17(4).

Ma, X., Tao, Z., Wang, Y., Yu, H., Wang, Y., 2015. Long short-term memory neural network for traffic speed prediction using remote microwave sensor data. Transportation Research Part C, 54, 187-197.

Messer, C., & Thomas Urbanik, I. I., 1998. Short-term freeway traffic volume forecasting using radial basis function neural network. Transportation Research Record Journal of the Transportation Research Board, 1651(1), 39-47.

Moniz, J. R. A., Krueger, D., 2018. Nested LSTMs. Proceedings of Machine Learning Research, 77, 530-544.

Okutani, I., Stephanedes, Y.J., 1984. Dynamic prediction of traffic volume through Kalman filtering theory. Transport. Res. Part B 18 (1), 1–11.

Oquab, M., Bottou, L., Laptev, I., Sivic, J., 2014. Learning and Transferring Mid-level Image Representations Using Convolutional Neural Networks. Computer Vision and Pattern Recognition (pp.1717-1724). IEEE.

Park, D., Rilett, L. R., 2010. Forecasting freeway link travel times with a multilayer feed-forward neural network. Computer-Aided Civil and Infrastructure Engineering, 14(5), 357-367.

Sabour, S., Frosst, N., Hinton, G. E., 2017. Dynamic routing between capsules. In Advances in Neural Information Processing Systems, 3856-3866.

Smola, A. J., Schölkopf, B., 2004. A tutorial on support vector regression. Statistics and Computing, 14, 199–222.

Srivastava, N., Hinton, G., Krizhevsky, A., Sutskever, I., Salakhutdinov, R., 2014. Dropout: a simple way to prevent neural networks from overfitting. Journal of Machine Learning Research, 15(1), 1929-1958.

Tan, M. C., Wong, S. C., Xu, J. M., Guan, Z. R., Zhang, P., 2009. An aggregation approach to short-term traffic flow prediction. IEEE Transactions on Intelligent Transportation Systems, 10(1), 60-69.

Tieleman, T., Hinton, G. 2012. Lecture 6.5-rmsprop: Divide the gradient by a running average of its recent magnitude. COURSERA: Neural Networks for Machine Learning, 4, 26–31.

Van Lint, J.W.C., Hoogendoorn, S.P., Zuylen, H.V., 2005. Accurate freeway travel time prediction with state-space neural networks under missing data. Transportation Research Part C, 13(5), 347-369.

Voort, M. V. D., Dougherty, M., Watson, S., 1996. Combining Kohonen maps with ARIMA time series

models to forecast traffic flow. Transportation Research Part C, 4(5), 307-318.

Williams, B., 2001. Multivariate vehicular traffic flow prediction: evaluation of ARIMAX modeling. Transportation Research Record Journal of the Transportation Research Board, 1776(1), 194-200.

Williams, B. M., Hoel, L. A., 2003. Modeling and forecasting vehicular traffic flow as a seasonal ARIMA process: theoretical basis and empirical results. Journal of Transportation Engineering, 129(6), 664-672.

Williams, B., Durvasula, P., Brown, D., 1998. Urban freeway traffic flow prediction: application of seasonal autoregressive integrated moving average and exponential smoothing models. Transportation Research Record, 1644(1), 132-141.

Wu, C. H., Ho, J. M., Lee, D. T., 2004. Travel-time prediction with support vector regression. IEEE Transactions on Intelligent Transportation Systems, 5(4), 276-281.

Wu, Y., Tan, H., 2016. Short-term traffic flow forecasting with spatial-temporal correlation in a hybrid deep learning framework. ArXiv Preprint ArXiv: 1612.01022.

Yin, H., Wong, S. C., Xu, J., & Wong, C. K., 2002. Urban traffic flow prediction using a fuzzy-neural approach. Transportation Research Part C, 10(2), 85-98.

Yu, H., Wu, Z., Wang, S., Wang, Y., Ma, X., 2017. Spatiotemporal recurrent convolutional networks for traffic prediction in transportation networks. Sensors, 17(7).

Zhang, J., Zheng, Y., Qi, D., 2016. Deep Spatio-temporal residual networks for citywide crowd flows prediction. AAAI, 1655-1661.

Zhang, Y., Zhang, Y., Haghani, A., 2014. A hybrid short-term traffic flow forecasting method based on spectral analysis and statistical volatility model. Transportation Research Part C, 43(1), 65-78.

Zheng, Z., Su, D., 2014. Short-term traffic volume forecasting: a k-nearest neighbor approach enhanced by constrained linearly sewing principle component algorithm. Transportation Research Part C, 43, 143-157.

Zhu, J. Z., Cao, J. X., Zhu, Y., & Zhu, Y., 2014. Traffic volume forecasting based on radial basis function neural network with the consideration of traffic flows at the adjacent intersections. Transportation Research Part C, 47(2), 139-154.